\pdfoutput=1

\documentclass[11pt]{article}

\usepackage{ACL2023}

\usepackage{times}
\usepackage{latexsym}

\usepackage[T1]{fontenc}

\usepackage[utf8]{inputenc}

\usepackage{microtype}

\usepackage{inconsolata}

\usepackage{multicol}
\usepackage{multirow}

\usepackage{graphicx}
\usepackage{expex}
\usepackage{xspace}
\usepackage{xurl}

\newcommand{\trr}{AR\xspace}

\newcommand\fauxsc[1]{\fauxschelper#1 \relax\relax}
\def\fauxschelper#1 #2\relax{%
  \fauxschelphelp#1\relax\relax%
  \if\relax#2\relax\else\ \fauxschelper#2\relax\fi%
}
\def\Hscale{.85}\def\Vscale{.74}\def\Cscale{1.12}
\def\fauxschelphelp#1#2\relax{%
  \ifnum`#1>``\ifnum`#1<`\{\scalebox{\Hscale}[\Vscale]{\uppercase{#1}}\else%
    \scalebox{\Cscale}[1]{#1}\fi\else\scalebox{\Cscale}[1]{#1}\fi%
  \ifx\relax#2\relax\else\fauxschelphelp#2\relax\fi}

\newcommand{\gpttwodistilled}{{\textsc{GPT-2 bft bd}}\xspace}
\newcommand{\gptthreeprompt}{{\textsc{GPT-3 pe}}\xspace}
\newcommand{\cohereprompt}{{Cohere \textsc{pe}}\xspace}
\newcommand{\human}{{\small\textsc{human}}\xspace}

\title{The economic trade-offs of large language models:\\ A case study}

 \author{Kristen Howell, Gwen Christian, Pavel Fomitchov, Gitit Kehat, Julianne Marzulla, \\ \bf{Leanne Rolston, Jadin Tredup, Ilana Zimmerman, Ethan Selfridge \and Joseph Bradley} \\
         LivePerson Inc., Seattle, Washington, U.S.A\\
         \{khowell, gchristian, pfomitchov, jmarzulla, jtredup, \\izimmerman, eselfridge, jbradley\}@liveperson.com\\
         \{gitit.kehat, leannerolston\}@gmail.com}

\begin{document}
\maketitle
\begin{abstract}
Contacting customer service via chat is a common practice. Because employing customer service agents is expensive, many companies are turning to NLP that assists human agents by auto-generating responses that can be used directly or with modifications. Large Language Models (LLMs) are a natural fit for this use case; however, their efficacy must be balanced with the cost of training and serving them. This paper assesses the practical cost and impact of LLMs for the enterprise as a function of the usefulness of the responses that they generate. We present a cost framework for evaluating an NLP model's utility for this use case and apply it to a single brand as a case study in the context of an existing agent assistance product. We compare three strategies for specializing an LLM \---\ prompt engineering, fine-tuning, and knowledge distillation \---\ using feedback from the brand's customer service agents. We find that the usability of a model's responses can make up for a large difference in inference cost for our case study brand, and we extrapolate our findings to the broader enterprise space.
\end{abstract}

\section{Introduction}\label{sec:intro}
Amidst increased automation, human agents continue to play an important role in providing excellent customer service. While many conversations are automated in text-based customer support, others are routed to human agents who can handle certain customer concerns more effectively. Agents often handle multiple conversations at once, consulting customer account information and brand policies while maintaining these conversations. As agents are expensive to staff, many companies are seeking ways to make their work more efficient.

\begin{figure*}
    \centering
    \includegraphics[width=.4\linewidth]{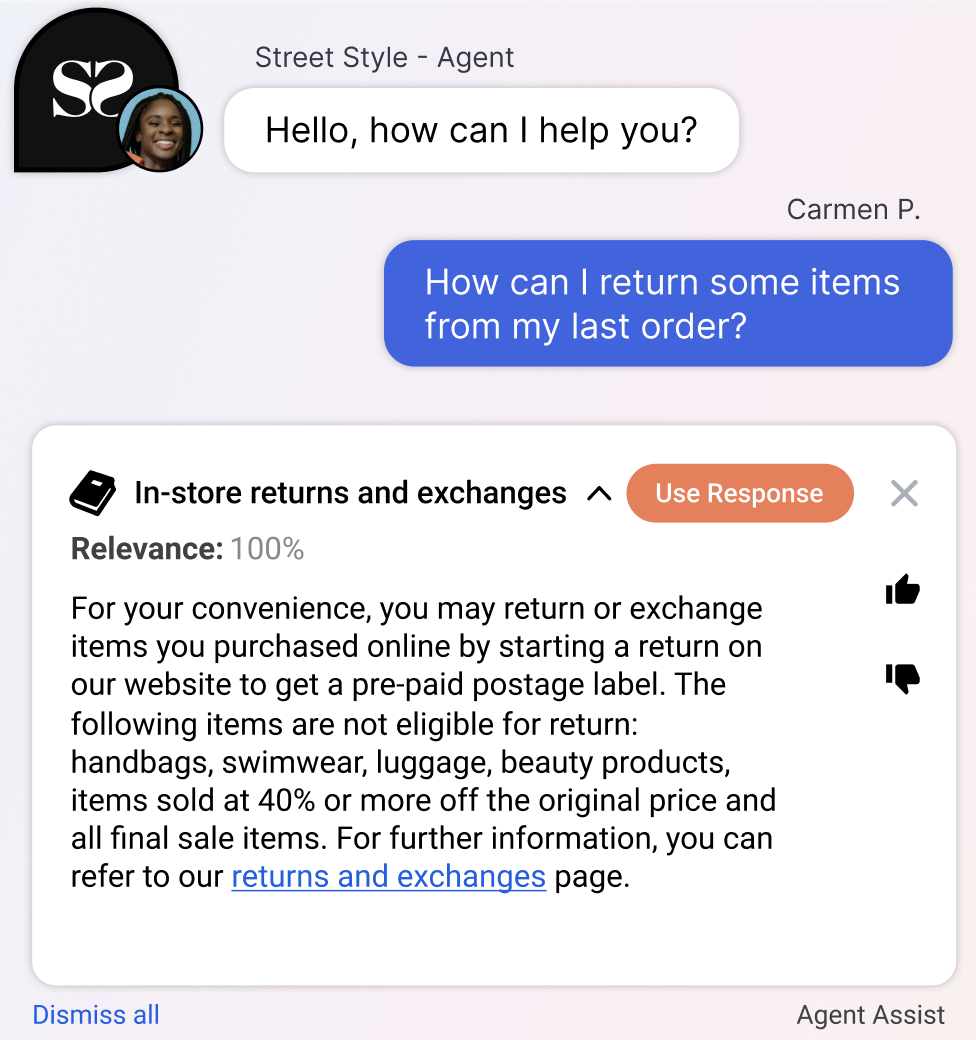} \hspace{.5cm}\includegraphics[width=.4\linewidth]{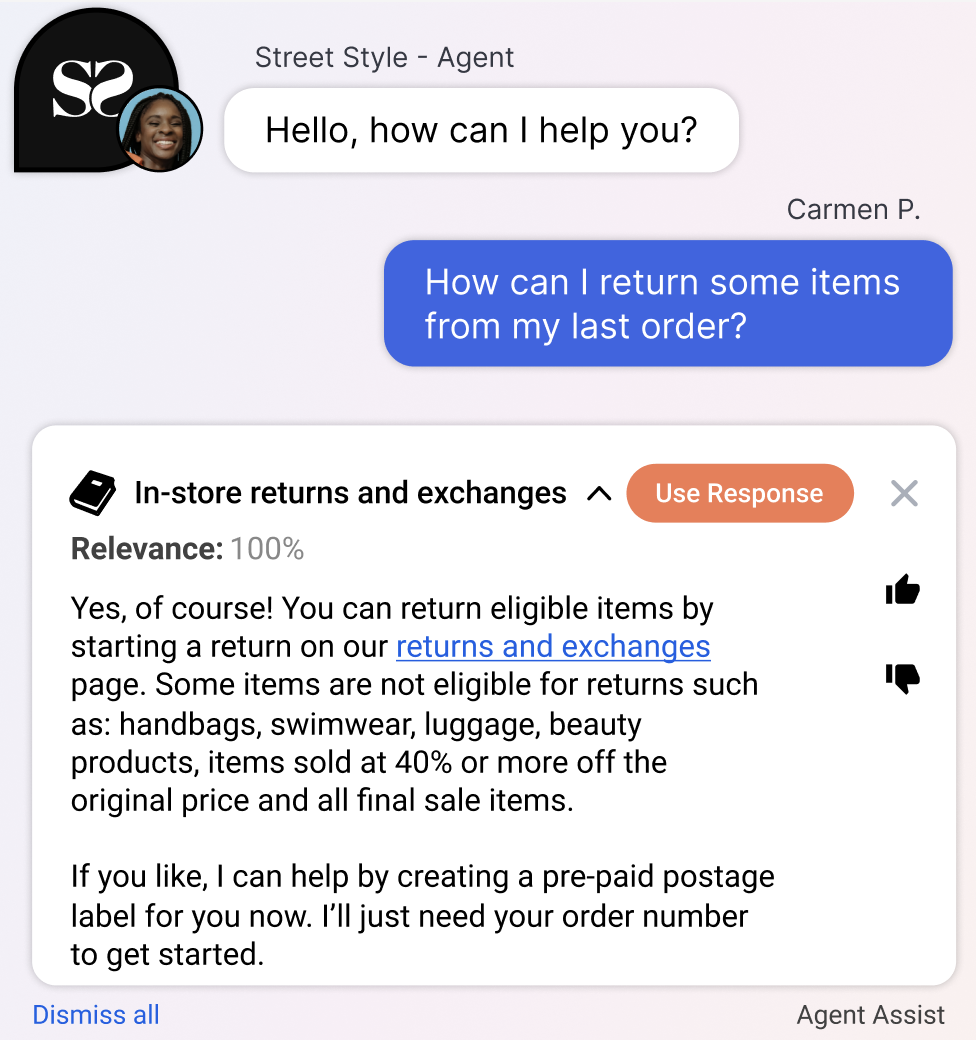}
    \caption{Conversation Assist as a system that returns canned responses (left), compared with the product described in this paper, which generates suggestions from LLMs (right).}
    \label{fig:CA}
\end{figure*}

LivePerson's Conversation Assist,\footnote{\url{https://developers.liveperson.com/conversation-assist-overview.html}} illustrated in Figure~\ref{fig:CA}, accelerates agents by automatically generating suggestions that the agent can either send, edit and then send, or ignore. Conversation Assist can both reduce agent response time and improve response quality, as a well-trained model may provide more consistent, higher quality responses than inexperienced agents or agents adversely impacted by external factors. These benefits lead to greater cost savings and increased customer satisfaction (CSAT) scores, not to mention providing a supervisory mechanism that is critical for brand control and model improvement.

Large Language Models (LLMs) are a natural fit for this technology, as they have achieved high performance on response generation tasks \citep[][inter alia]{adiwardana2020towards,hosseini2020simple,zhang-etal-2020-dialogpt}, but they are expensive to train and serve. For example, the inference cost for each response using a distilled GPT-2 model and an Nvidia A100 GPU is \textcent.0011,\footnote{We found the Nvidia A100 GPU to be the most inexpensive option, with an Nvidia V100 GPU costing \textcent0.0019} while the inference cost using the GPT-3-based Davinci model through OpenAI's API is \textcent1.10 \cite{openaipricing}.\footnote{Assuming a context and response length of 550 tokens.} 

LLM economics and enterprise applications are highly fluid. First, individual partnership agreements may differ from the published API cost, and the rapid pace of innovation in the space will necessarily impact the cost of training and serving these models. Second, as brands vary widely, a useful agent assistance model must be customized to the brand's use case and performance requirements. We propose a simple and flexible cost framework that can be applied to various LLM and brand scenarios. This framework, Expected Net Cost Savings (ENCS), combines the probability and cost savings of an agent accepting or editing a response with the cost of generating the response. ENCS can be applied at the message level or in the aggregate.

With one brand as a case study, we explore ENCS with various methods of model customization. Using feedback from the brand's customer service agents, we evaluated fine-tuning, prompt engineering, and distillation to adapt and optimize GPT-2 \citep{radford2019language}, GPT-3 \citep{brown2020language,openaigpt3}, and Cohere \citep{coheregeneration}. These strategies can lead to an agent usage rate of 83\% (including both direct use and editing) and an annual cost savings of \$60,000 for our case-study brand \---\ 60\% of their total agent budget.

We generalize this case study to a broader range of brands and models. We find that low perplexity correlates with the probability that an agent will use a response, and we extrapolate from this finding to use perplexity to estimate the ENCS for additional model customization strategies. We apply ENCS to each configuration, and while models, prices, and use cases will change over time, we expect that this framework can be continuously leveraged for decision making as technology evolves.

\section{Related Work}\label{sec:related-work}
Transformers \citep{vaswani2017attention} have dominated response generation tasks: DialogGPT \citep{zhang-etal-2020-dialogpt}, Meena \citep{adiwardana2020towards}, \textsc{soloist} \citep{peng2021soloist}, BlenderBot \citep{roller-etal-2021-recipes}, PLATO-XL \citep{bao2022plato}, LaMDA \citep{thoppilan2022lamda}, \textsc{godel} \citep{peng2022godel}. Each of these approaches fine-tunes a large pre-trained LM to task-oriented dialog or chit chat using curated dialogs. In some cases, additional tasks, such as the discriminative training tasks of \citealt{thoppilan2022lamda}, are also used. When data is not available for fine-tuning, prompting with a single example has proven quite effective \citep{min2022rethinking}, and for large enough models, prompting that demonstrates breaking tasks into discrete components \citep{wei2022chain} has performed on par with fine-tuned models \citep{chowdhery2022palm}. 

The size of these LLMs plays a significant role in their high performance \citep{chowdhery2022palm}, but in a deployed setting, this size can be quite costly. Quantization \citep{whittaker2001quantization, shen2020q}, pruning \citep{han2015learning,han2016eie} and knowledge distillation \citep{hinton2015distilling,sanh2019distilbert} are common strategies for size reduction with minimal impact to performance. Here we focus specifically on distillation using a language modeling task to reduce model size while simultaneously adapting the model to the data following \citet{ryu2020knowledge} and \citet{howell2022domain}.

 Response generation is difficult to evaluate holistically. Some have focused on relevance and level of detail \citep{zhang-etal-2020-dialogpt,adiwardana2020towards,thoppilan2022lamda}, humanness \citep{zhang-etal-2020-dialogpt,roller-etal-2021-recipes} and overall coherence or interestingness \citep{bao2022plato,thoppilan2022lamda}. In contrast, we follow \citet{thoppilan2022lamda} and \citet{peng2022godel} who consider helpfulness and usefulness as broader measures of response quality, but we ground these judgements in the customer service use case by having real agents judge the usefulness of model outputs.

\section{Expected Net Cost Savings (ENCS)}\label{sec:cost-savings}
ENCS combines model performance, model cost, and agent cost: If an agent saves time by using a model's response, then there is a cost savings. More formally, ENCS is defined as the probability that a response is used ($P(U)$) multiplied by the savings in dollars for each used response ($S_{U}$), less the cost of generating that response ($C$), as in (\ref{ex:ecs-simple}).

\ex\label{ex:ecs-simple}
$ENCS = P(U) * S_{U} - C$
\xe

\noindent Because agents are not limited to simply using a response as-is but may also choose to edit the response or ignore it altogether, equation (\ref{ex:ecs-simple}) may be modified to account for the probability and savings associated with editing ($P(E)$ and $S_{E}$) or the cost of ignoring ($P(I)$ and $S_{I}$)\footnote{In most cases, $S_{I}$ is a negative number, as reading a response and choosing not to use it would cost time and money.} as well:

\ex\label{ex:ecs-edit}
$ENCS = ((P(U) * S_{U}) + (P(E) * S_{E}) + (P(I) * S_{I}) - C$
\xe

We can estimate $S$ from the agent's hourly rate ($R$), the average time it takes for agents to respond to a message without Conversation Assist ($T_{r}$), and the amount of time an agent spends for each accepted, edited, or ignored message ($T_{x}$). 

\begin{figure}
    \centering 
    \includegraphics[width=0.9\linewidth]{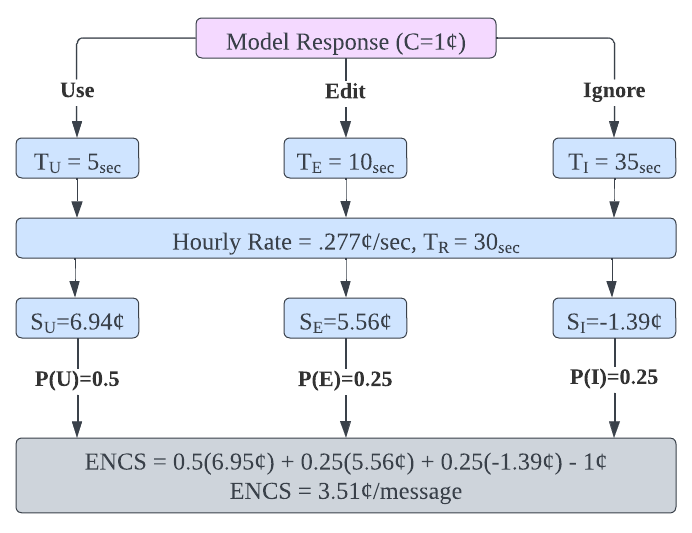}
    \caption{A toy example of an ENCS calculation.}
    \label{fig:ENCS}
\end{figure}

\ex\label{ex:savings-per-selection}
$S_{x} = R(T_{r} - T_{x})$
\xe
Figure \ref{fig:ENCS} provides a toy example of this calculation.

\subsection{Simplifying Assumptions}\label{sec:assumptions}
This model makes a number of simplifying assumptions. We assume that agents always have conversations to respond to or some other work to do. We exclude the problem of workforce optimization from our framework, noting that when fewer agents are needed to handle the conversational traffic, workforce can be reduced. We also exclude R\&D cost, but return to this factor in section~\ref{sec:beyond}.

Furthermore, we omit any discussion of the cost of an agent using an inappropriate or factually incorrect response. For the purposes of this model, we assume that agents read all suggestions carefully, but a deeper analysis of the risk and cost of these errors is a critical area for further study.

\section{Case Study}\label{sec:case-study}
We focus on a single brand to evaluate the use of LLMs for Conversation Assist and explore the application of ENCS for making product decisions. We evaluate three model customization strategies using manual ratings from brand agents. We then evaluate how well these ratings relate to perplexity and use this to assess a larger set of models. Finally, we estimate ENCS and discuss the implications.

\subsection{Case Study Brand}\label{sec:brand}
We partnered with a single brand, who we will refer to as Anonymous Retailer (\trr), for this case study. \trr's customer base includes both consumers and sellers who consign items through \trr's platform. Because \trr's agents are trained across different customer concern categories, they can provide expert feedback on a wide range of data. 

At the time of writing, \trr\ has about 350 human agents who use LivePerson's chat platform. \trr\ supports about 15,000 conversations per month, and uses chat bots for simple tasks and routing, while their human agents send 100,000 messages per month on average. 
In comparison, the average number of conversations per month for brands on LivePerson's platform is 34,000, with a median of 900 monthly conversations per brand and a standard deviation of 160.

\begin{table*}[]
    \centering
    \small
    \resizebox{\textwidth}{!}{
    \begin{tabular}{l||lrr|lrr|lrr}
    & \multicolumn{3}{c}{\textbf{Fine-tuning}} & \multicolumn{3}{c}{\textbf{Distillation}} & \multicolumn{3}{c}{\textbf{2nd Fine-tuning}}\\
\textbf{Model Name}	& \textbf{Dataset}	& \textbf{\# Convs}	&	\textbf{\# Steps}	& \textbf{Dataset}	& \textbf{\# Convs}	&	\textbf{\# Steps}	& \textbf{Dataset}	& \textbf{\# Convs}	&	\textbf{\# Steps}	\\\hline\hline
\textsc{GPT-2 bft bd}$^*$	&	brand	&	100,059	&	15,000	&	brand	&	100,059	&	67,014	&		&		&		\\
Cohere \textsc{pe}$^*$	&		&		&		&		&		&		&		&		&		\\
\textsc{GPT-3 pe}$^*$	&		&		&		&		&		&		&		&		&		\\
\textsc{GPT-2 bft}	&	brand	&	100,059	&	15,000	&		&		&		&		&		&		\\
\textsc{GPT-2 bft bf bft}	&	brand	&	100,059	&	34,000	&	brand	&	100,059	&	67,014	&	brand	&	100,059	&	15,000	\\
\textsc{GPT-2 gft bd bft}	&	general	&	236,769	&	34,000	&	brand	&	100,059	&	67,014	&	brand	&	100,059	&	28,000	\\
\textsc{GPT-2 gft gd bft}	&	general	&	236,769	&	34,000	&	general	&	236,769	&	1,264,352	&	brand	&	100,059	&	28,000	\\
\textsc{GPT-2}	&		&		&		&		&		&		&		&		&		\\
\textsc{GPT-2 XL gft gd bft}	&	general	&	236,769	&	120,000	&	general	&	236,769	&	1,264,352	&	brand	&	100,059	&		\\
Cohere \textsc{ft}	&	brand	&	50	&		&		&		&		&		&		&		\\
\textsc{GPT-3 bft}	&	brand	&	50	&	4 epochs	&		&		&		&		&		&		\\
    \end{tabular}}
    \caption{Model adaptation configurations. $^*$ indicates that this model's outputs were manually evaluated. \\\textit{\fauxsc{bft} = fine-tuned on \trr\ brand data, \fauxsc{gft} = fine-tuned on the general dataset, \textsc{bd} = distilled using \trr\ brand data, \fauxsc{gd} = distilled using the general dataset, \fauxsc{pe} = prompt engineered.}}
    \label{tab:all-models}
\end{table*}

\subsection{Data sets}\label{sec:data}
We constructed three datasets: brand-specific training, brand-specific test, and general training. We de-identified data, replacing each entity with a random replacement. For the test set, we manually ensured that the de-identification was internally consistent across the conversation for agent and consumer names, addresses, and order numbers. 

The brand-specific data comprises English customer service conversations from 2022 that include human agent and bot messages. We filtered these conversations to ensure that they had at least two agent turns, more human agent than bot messages, and a positive Meaningful Conversation Score.\footnote{For more information on Meaningful Conversation Score, see: \url{https://knowledge.liveperson.com/data-reporting-meaningful-conversation-score-(mcs)-meaningful-conversation-score-(mcs)-overview.html/}}

From this filtered data, we randomly sampled 100,059 conversations to make up our training set. From the remainder, we curated a brand-specific test set by manually selecting 287 conversations where the customer's goal could be clearly established from the context of the conversation. We constructed the general training set from five additional retail brands whose product lines fall into similar categories as \trr. We filtered and processed the data using the method described above and selected 70,000 conversations per brand, or used the entirety of the brand's data if there were fewer than 70,000 conversations. The total size of the general training set is 236,769 conversations. For more details on these datasets, see Appendix \ref{appendix:dataset}.

\subsection{Model Customization}\label{sec:models}

We explored three standard model customization strategies: prompt engineering, fine-tuning, and knowledge distillation. Using these strategies, we tested eleven configurations (Table~\ref{tab:all-models}). We evaluated three of these configurations with the judgements of \trr\ agents, and for the remainder we extrapolated usability scores from the model's perplexity over the test set. 

\subsubsection{Prompt Engineering}

\begin{figure*}
    \centering
    \includegraphics[width=.9\linewidth]{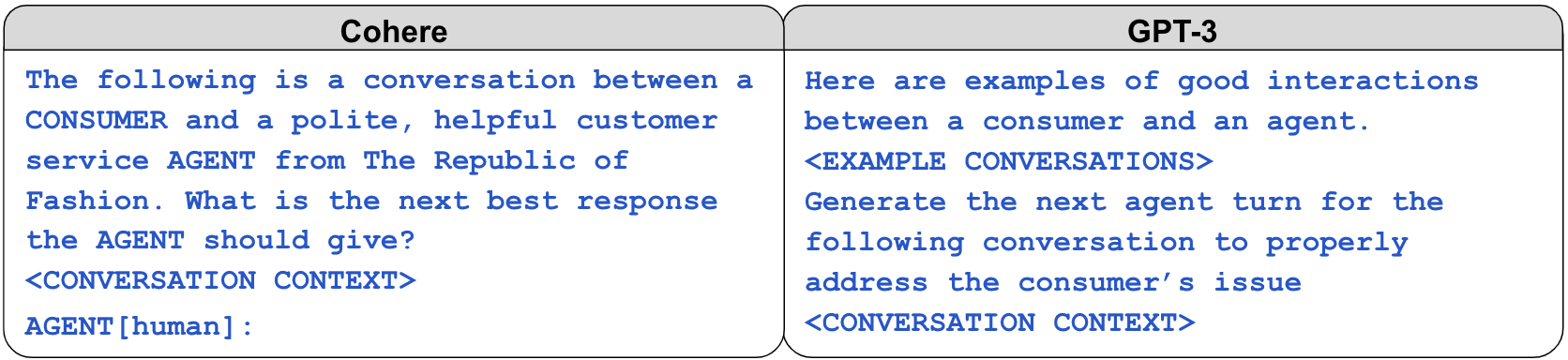}
    \caption{Prompts used for Cohere and GPT-3\footnotemark}
     \label{fig:prompts}
\end{figure*}

\paragraph{GPT-3}
We prompted the \textsf{text-davinci-003} GPT-3 model \citep{openaigpt3}, following OpenAI's best practices for prompt engineering \citep{openaiprompteng}. After some experimentation, we found that the most effective prompt for our use case (Figure~\ref{fig:prompts}) used a hand-constructed exemplar conversation and explicitly instructed the model to generate a response that would address the consumer's issue.

\paragraph{Cohere}
Following Cohere's best practices \cite{cohereprompt}, we tested both verbose and concise prompts with the XLarge Cohere model \cite{coheregeneration}. Unlike GPT-3, we found that using a prompt without an exemplar conversation (Figure~\ref{fig:prompts}) resulted in better performance. 

\subsubsection{Fine-Tuning}\label{sec:fine-tuning}
\paragraph{GPT-2} We fine-tuned GPT-2 \citep{radford2019language} using a language modeling task over conversational data on either the brand-specific dataset or the general dataset described in section~\ref{sec:data}. We started with a learning rate of 0.00008 with a linear scheduler and no warm up steps and trained until perplexity plateaued. 

\paragraph{GPT-3} We fine-tuned the \textsf{text-davinci-003} GPT-3 model from OpenAI on a conversational prompt-completion task using instructions and an exemplar conversation as the prompt 
and the human-agent response as the output. The dataset consisted of 50 random examples from the brand-specific training set.

\paragraph{Cohere} We fine-tuned Cohere's XLarge model with Cohere's API \cite{coherecustom} and a random subset of 50 conversations from the brand-specific dataset. We tested verbose and concise prompts as well as \textsc{eos} token placement, and found that a shorter prompt with an \textsc{eos} token after each turn worked best.

\subsubsection{Distillation}\label{sec:distillation}
To reduce latency and cost to serve by almost half, we distilled our fine-tuned GPT-2 models using the Transformers library \cite{transformersdistil}, following the method set forth by \citet{sanh2019distilbert} and the language modeling training task of \citet{radford2019language}. For distillation, we used either the brand-specific or the general dataset. We started with a learning rate of 0.0005 using a linear scheduler and trained for 3 epochs. Because the OpenAI and Cohere API's do not make the logits of the whole vocabulary available at inference, we are unable to distill these models using \citeauthor{sanh2019distilbert}'s methodology.

\footnotetext{This example prompt uses a fictitious brand name for anonymity.}

\subsection{Metrics and Results}\label{sec:tasks}

\subsubsection{Response Usability}\label{sec:ru-task}
While previous work has assessed the helpfulness or usability of a response with crowd-sourced judgments \citep{thoppilan2022lamda,peng2022godel}, we worked with nine agents at \trr\ who already use our Conversation Assist product. For each conversation and suggested response, we asked them whether they would use the suggested response as-is; edit it to change specific details, add to it, or remove parts of it; or ignore the suggestion altogether. The full annotation instructions are given in Appendix~\ref{appendix:ru}.

\begin{table}[]
    \centering
    \small
    \begin{tabular}{l|ccc}
 \textbf{Model Name} &  \textbf{\%Ignore} &  \textbf{\%Edit} &  \textbf{\%Use}\\\hline\hline
\human &  10 &  12 &  77\\
\gpttwodistilled\ &  28 &  16 &  57\\
\cohereprompt\ &  22 &  20 &  58\\
\gptthreeprompt\ &  \bf 17 & 14 & \bf 69\\
    \end{tabular}
    \caption{The percentage of responses that agents said they would use, edit, or ignore. Five agents annotated each conversation, judgements are counted individually.}
    \label{tab:usability-results}
\end{table}

Table~\ref{tab:usability-results} shows annotated Response Usability (RU) scores for three models. Even when shown the response that an \trr\ agent had actually used in the conversation (\human), agents said that they would use this response only 77\% of the time and would ignore it 10\% of the time. This indicates a high level of personal preference among the agents, and sets a noteworthy upper limit on the usability we could expect from model outputs. Agents said that they would use the \gptthreeprompt\ suggestion 69\% of the time compared with \gpttwodistilled and \cohereprompt\ at only 57\% and 58\%, respectively. 

As the use rate increases, the edit rate and ignore rates both decrease, indicating that conversations resulting in editable prompts for some models can result in usable prompts for another model. We also note, that while the use rate was similar for \gpttwodistilled\ and \cohereprompt, the edit rate was much higher for cohere, highlighting the importance of assessing the cost savings of an editable response vs. ignoring the response entirely.  

We also annotated these conversations for the Foundation Metrics in \citealt{thoppilan2022lamda} and found a correlation between responses that were sensible, specific and role-consistent and those that the agents said they would use. Detailed analysis of these labels and their correlation are in Appendix~\ref{appendix:sup-analysis}. This additional annotation revealed that, of the three models, \gpttwodistilled was most likely to generate a consumer turn rather than an agent turn or to generate a turn that was not relevant to the conversation, which may account for its high ignore rate. We also note that virtually all responses generated by the three models were labeled `safe' by the annotators.

\subsubsection{Perplexity}\label{sec:perplexity-task}
\citet{adiwardana2020towards} found that sensibleness and specificity corresponded with the model's perplexity, inspiring us to use perplexity to extrapolate our manual evaluation of three models to a broader set of model configurations. After reproducing \citeauthor{adiwardana2020towards}'s finding for sensibleness and specificity using our data (see Appendix~\ref{appendix:linear_modeling}), we investigated the correlation between perplexity and response usability. For each conversation context in the evaluation set, we calculate the perplexity for the generated response for each LLM using the average log likelihood of each token, following equation (\ref{ex:perplexity}). 

\ex\label{ex:perplexity}
$PP(W) = \sqrt[N]{\frac{1}{P(w_1,w_2,...,w_N}})$
\xe

\setlength{\tabcolsep}{1mm}
\begin{table}
    \centering
    \small
    \begin{tabular}{l|c|ccc}
\textbf{Model Name}	&	\textbf{PPL}	&	\textbf{\%Ignore}	&	\textbf{\%Edit}	&	\textbf{\%Use}	\\\hline\hline
\textsc{GPT-2 bft}	&	4.27	&	20.8	&	16.8	&	62.4	\\
\textsc{GPT-2 bft bd bft}	&	4.50	&	21.0	&	16.9	&	62.1	\\
\textsc{GPT-2 gft bd bft}	&	4.05	&	20.7	&	16.7	&	62.6	\\
\textsc{GPT-2 gft gd bft}	&	4.15	&	20.7	&	16.8	&	62.5	\\
\textsc{GPT-2}	&	7.08	&   22.7	&	17.8	&	59.5	\\
\textsc{GPT-2 XL gft gd bft}	&	5.31	&	21.5	&	17.2	&	61.3	\\
Cohere \textsc{ft}	&	1.93	&	19.3	&	16.0	&	64.7	\\
\textsc{GPT-3 bft}	&	4.14	&	20.7	&	16.8	&	62.5	\\
    \end{tabular}
    \caption{Average perplexity (PPL)\footnotemark and projected Response Usability (RU) scores. See Table~\ref{tab:all-models} for descriptions and naming conventions for the models.}
    \label{tab:pplx}
\end{table}
\footnotetext{On the rare occasion that a model did not generate a response, we exclude that data point from the average perplexity as it would heavily skew the average.}

Using all annotated LLMs' suggested responses across all conversations in the evaluation set, we fit a set of linear regression models using the perplexity of the generated agent turn as our independent variable, and the probability of use, edit, and ignore as our dependent variables. Individual linear models trained on the output of a single LLM did not show statistical significance; however, models trained on the output of all LLMs did show significance in the F-statistic (p < 0.05 for P(edit), p < 0.001 for P(use) and P(ignore)). Extrapolating from these linear models allows us to illustrate potential cost savings for more models than we were able to annotate. These linear models predict the RU scores in Table~\ref{tab:pplx}.

\subsubsection{Expected Net Cost Savings (ENCS)}\label{sec:encs-task}
\begin{table}
    \centering
    \small
    \begin{tabular}{l|ccc}
        \textbf{Model Name} &  \textbf{\textsc{encs}/message} & \textbf{\textsc{encs}/year} \\\hline\hline
         \gpttwodistilled  & \textcent 4.47 & \$53,653 \\
        \cohereprompt & \textcent 4.58 & \$55,000  \\
        \gptthreeprompt & \textcent 4.24 & \$50,920  \\\hline
        \textsc{GPT-2 bft} & \textcent 4.97 & \$59,687  \\
        \textsc{GPT-2 bft bf bft} & \textcent 4.96 & \$59,527  \\
        \textsc{GPT-2 gft bd bft} & \textcent 4.99 & \$59,851  \\
        \textsc{GPT-2 gft gd bft} & \textcent 4.98 & \$59,786  \\
        \textsc{GPT-2} & \textcent 4.81 & \$57,668  \\
        \textsc{GPT-2XL gft gd bft} & \textcent 4.90 & \$58,802  \\
        Cohere \textsc{ft} & \textcent 4.62 & \$55,391  \\
        \textsc{GPT-3 bft} & -\textcent 1.56 & -\$18,691  \\
        
    \end{tabular}
    \caption{\trr's estimated cost savings per model using equation~\ref{ex:ecs-edit} and the usage rates in Table~\ref{tab:usability-results}. For models below the line, we we use extrapolated usage rates using perplexity from Table~\ref{tab:pplx}. The assumptions used to calculate the ENCS are described in section~\ref{sec:encs-task}. See Table~\ref{tab:all-models} for descriptions and naming conventions for these models.}
    \label{tab:cost-savings-per-model}
\end{table}

We calculate the ENCS for each model using equation (\ref{ex:ecs-edit}), repeated here in (\ref{ex:ecs-edit-2}). 

\ex\label{ex:ecs-edit-2}
$ENCS = ((P(U) * S_{U}) + (P(E) * S_{E}) + (P(I) * S_{I}) - C$
\xe

\noindent $P(U)$, $P(E)$, and $P(I)$
are the frequency with which the LLM's response was accepted, edited, or ignored in the test set. $S_U$, $S_E$, and $S_I$ are calculated assuming that an agent costs \$10.00 per hour and averages 30 seconds per message without Conversation Assist. With Conversation Assist, we assume that the agent saves 25 seconds for each accepted response, 20 seconds for each edited response and spends an extra 5 seconds for each ignored response. We also assume that each response costs \textcent0.002 to generate for a GPT-2 model, \textcent0.0011 for a distilled GPT-2 model, \textcent 1.09 for the base model and \textcent 6.54 for a fine-tuned model through OpenAI's API and \textcent 0.25 for the base model and \textcent 0.50 for a fine-tuned model through Cohere's API.\footnote{We estimate GPT-2's cost based on a latency of 19.57 milliseconds per inference for the full-sized model and 11.60 ms for the distilled model, and a cost of \$3.53 per hour renting an Nvidia A100 GPU from GCP for 8 hours a day. OpenAI and Cohere's API costs are come from \citealt{openaipricing} and \citealt{coherepricing} at the time of writing.}

Using the RU scores in Tables~\ref{tab:usability-results} and~\ref{tab:pplx}, we estimate that \trr's cost savings per message would be \textcent 4.47 using the \gpttwodistilled\ model compared with \textcent 4.24 using \gptthreeprompt, as detailed in Table~\ref{tab:cost-savings-per-model}. ENCS per year is calculated based on \trr's annual agent message volume of 1,200,000.  

The factor with the largest impact on \trr's cost savings is  the usefulness of the predictions, as the best annotated model (\gptthreeprompt)'s predictions are used or edited only 5\% more often than the fastest (\gpttwodistilled), while its cost was almost 100 times higher (\textcent1.09 vs \textcent0.0011). Despite this, the difference in ENCS between these two models is minimal and only amounts to about \$3k per year. In general, the RU and ENCS are higher for the extrapolated results, which are somewhat less reliable, but they lead to one important insight: in this case, the inference cost for a fine-tuned GPT-3 model is too high for the customer to realize savings.

\section{Beyond a single case study}\label{sec:beyond}
\begin{figure}
    \centering
    \includegraphics[width=\linewidth]{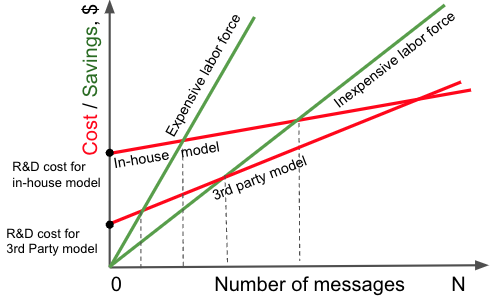}
    \caption{Factors impacting when a brand will break even when using an agent assistance model.}
    \label{fig:break-even}
\end{figure}

To decide which of these models will lead to the greatest ROI for a brand, we must consider the break-even point for each model based on the ENCS (which includes agent labor and model inference costs) as well as R\&D cost and message volume. This can be visualized with Figure~\ref{fig:break-even}, which shows that ROI is reached when the amount that labor cost is offset (green) intersects with the amount that has been spent on the model (red). The number of suggestions needed to break even ($N_r$) is calculated with equation (\ref{ex:breakeven}), using the R\&D cost ($C_{R\&D}$), ENCS, and the cost to update and maintain the model (expressed as an average per message over time as $C_{m}$).

\ex\label{ex:breakeven}
$N_{r} = \frac{C_{R\&D}}{(ENCS - C_{m})}$
\xe

Given that the difference in ENCS per message across the models explored in this paper is not large, low R\&D cost is the main consideration to reach the fastest ROI. For a small brand sending 500,000 agent messages per year and saving about \$24,000 per year with any of the models, reducing the upfront R\&D cost would be critical. On the other hand, a large enterprise brand who will save \$950,000 per year over 20 million messages, will break-even on any R\&D cost fairly quickly. As a model with lower inference cost will offset high R\&D cost more quickly and lead to more savings over a longer period of time, inference cost is a much more important factor for a brand with high traffic. In Appendix~\ref{appendix:cost-savings}, we provide a detailed example of the impacts of these costs.

It is also worth noting that when choosing between in-house and third-party models, the difference in R\&D and maintenance cost may not be as significant as one might expect. While an in-house model requires up-front investment to train and serve, OpenAI and Cohere's LLMs at the time of writing require a fair amount of effort to prompt engineer for the best performance and these prompts should be customized to some degree for different brands and scenarios. From a maintenance perspective, we similarly find that while an in-house model must be refreshed, prompts must also be redesigned as third-party providers update and release new models.

Brands might also wish to consider factors that are not accounted for in this framework. Some brands would prefer to use an in-house model so that they can retain control over their data and protect their customer privacy by limiting access of their data to third-party vendors. An in-house model also provides more control over the model's suggestions, as well as control over when the model is updated or deprecated. Especially as technology develops, models become less expensive to train, and the performance of open-source models improves, these factors may carry even more weight.

\section{Conclusion}
In this case study, we demonstrated the utility of LLMs for agent assistance products, exploring 3 model adaptation strategies across 11 model configurations. Based on feedback from real customer service agents, we found that bigger is not always better, as the distilled GPT-2 model resulted in greater cost-savings than GPT-3, despite lower quality responses, because, at the time of writing, its inference cost is so much lower. These results empower near-term decision-making for integrating models like these into production.

However, with the rapidly shifting NLP landscape, a framework to assess the cost benefits of new technologies is critical to facilitate decisions about integrating them into products. The flexible framework presented in this paper, ENCS, enables NLP practitioners to invest in innovations that lead to tangible business benefits. We found that for this product, the impact of model quality far outweighs inference cost, pointing to the importance of continuing to push the state of the art, while considering practical expense. This framework empowers the NLP community to invest in the most cost-effective technology for their specific needs, even as that technology, its capability, and its pricing evolve.

\section*{Ethics Statement}\label{sec:ethics}
To protect customer and agent privacy, the data used to train and evaluate models was fully anonymized by replacing all customer or agent names, addresses, phone numbers, or other personal identifiers with a random name or string. We also compensated agents for annotations in line with their standard rate as agents at \trr.

While the tools described in this paper have the explicit goal of making agents' jobs easier, they - and specifically the lens of a cost savings analysis - have the potential to be used to motivate reductions in workforce, and we acknowledge the impact that this can have on the agents themselves. We also note that these tools can also improve the customer experience by reducing wait times, which can lead to fewer frustrated customers when they do interact with agents.

\section*{Limitations}\label{sec:limitations}
In this study, we collected feedback on the usefulness of model responses from customer service agents at \trr. These agents were recommended based on their availability and experience with Conversation Assist; however, we did not receive details about the agents such as their level of training or experience, which may have an impact on their preferences using the suggested responses. Furthermore, while agents in our study received a flat rate per judgment with no bonus or penalties to how they judged the response, some businesses have existing agent metrics (e.g.\ actual handle time, AHT targets, etc.) that could incentivize the agents to behave differently while performing their jobs. These metrics have the potential to exert pressure on agents in real-life situations to accept responses at a higher rate than in this study.

The linear models in section \ref{sec:perplexity-task} are based on the judgments of 5 agents on 3 LMM model outputs for 287 conversations. While they have shown a statistically significant relationship between usage rates and perplexity, this is a small pilot analysis. Additional data will be necessary to determine how well this generalizes.

Our cost savings framework also makes a number of simplifying assumptions about workforce optimization. We've noted some of these assumptions in section~\ref{sec:assumptions}, and they should be considered when leveraging this framework for different types of products. In addition, while the explicit goal of these models is to make agents' jobs easier, we expect from previous work studying vigilance tasks \citep{warm2008vigilance} that there can be an upper bound to how much cost could be saved with an excellent LLM, as there would be less benefit from the agent acting as a human in the loop as their vigilance wanes.

\section*{Acknowledgements}
We want to thank the customer service agents at \trr\ who provided judgements on the usability of responses, as well as Anna Folinsky and Daniel Gilliam who annotated the data for Foundation Metrics. We also appreciate engineering assistance that we received from Larry Cheng, Tyson Chihaya, Sid Naik, and Fazlul Shahriar.

\bibliography{paper}
\bibliographystyle{acl_natbib}

\appendix

\section{Training Details: Model Fine-tuning}\label{appendix:tuning}
\paragraph{GPT-2}

We fine-tuned the pretrained GPT-2 model from huggingface using either the brand-specific \trr\ or general dataset. Each training example has an end-of-text token appended to the beginning and end of the conversation and is padded with an added pad token. The resulting model has 117M parameters and a vocabulary size of 50258 (GPT-2 vocab size with an additional pad token). We started with a learning rate of 0.00008 with a linear scheduler and no warm up steps. The model was trained for 34000 steps across 4 Nvidia Tesla V100 GPUs, which equates to roughly 3 epochs for the \trr\ dataset and 5 epochs for the general dataset.

\paragraph{GPT-3}

We fine-tuned GPT-3 with prompt-completion pairs using the OpenAI API. We trained for 4 epochs using a total of 50 examples that were selected and split at random human-agent turns to append the preceding conversation to the prompt and the human-agent turn as the completion. Additionally, the prompt included a brief summary of the context before giving the conversational context, which includes a separator sequence to delineate the summary and the conversation. An example of a prompt-completion pair is given below:\\

\small
\fbox{\begin{minipage}{0.4\textwidth}
\textit{\textbf{Prompt:}}
\begin{quote}
    Summary: The following is a conversation between a CONSUMER and a polite, helpful, customer service AGENT from <BRAND\_NAME>.  

    CONSUMER: <consumer\_turn>
    
    AGENT[non-human]: <agent\_turn>

    \textit{...}

    AGENT[human]:
\end{quote}
\par \textit{\textbf{Completion:}}
\begin{quote}
    <agent\_response>
\end{quote}
\end{minipage}}
\normalsize
\paragraph{Cohere}

To fine-tune the Cohere model, we experimented with different configurations for pre-processing the input data that varied the input prompts and whether or not to use an end-of-sequence token between conversation turns. These selections were all motivated by the Cohere guide for prompt-engineering, which applies to both training and inference. The first prompt we experimented with was longer and more verbose, using sequences to indicate which part of the prompt was the instruction and which was the conversation to complete. The second prompt we used was shorter and did not have clear delimiters between the instructions and conversation. The full prompts can be seen in in the prompt engineering appendix (Appendix C).

\section{Training Details: GPT-2 Distillation}\label{appendix:distillation}
\paragraph{GPT-2}
We distilled our fine-tuned GPT-2 models using the distillation code provided by Huggingface. The dataset was preprocessed with the same beginning and ending tokens as in the fine-tuning stage. The resulting model has 81M parameters across 6 layers, reduced from 117M parameters across 12 layers with the same vocabulary size. Training started with a learning rate of 0.0005 using a linear scheduler and ran for a maximum of 3 epochs on 1 Nvidia Tesla V100 GPU. This resulted in 67,014 and 164,352 steps for distilling on the \trr\ and general datasets, respectively.

\section{Prompt Engineering}\label{appendix:prompt_engineering}

\paragraph{GPT-3}
\label{appendix:prompt_engineering_gpt3}

We experimented with several prompts before choosing one that gave adequate results without consuming too much of the token limit.  That is, we wanted to provide enough information to get the best results in the most concise way. 

First, we varied the verbosity of the framing of the request, changing factors such as whether the brand name was provided or whether there was a description of the product line:\\

\fbox{\begin{minipage}{0.42\textwidth}
\small
You are a customer service representative
\newline\newline
You are a customer service representative for a retail and consignment brand
\newline\newline
You are a customer service representative for a luxury retail and consignment brand
\newline\newline
You are a customer service representative for a luxury retail and consignment brand called The Republic of Fashion
\newline\newline
You are a customer service representative for a luxury retail and consignment brand called The Republic of Fashion which is an anonymized version of \trr.

\end{minipage}}
\\
\\
The quality of the responses did not vary based on the amount of detail given here, nor did they change when this was omitted, so we chose to omit it.

The next aspect we varied was the amount of detail given in the description of the examples:\\

\fbox{\begin{minipage}{0.42\textwidth}
\small
    Here are examples of good interactions
    \\\\
    Here are examples of good interactions between a consumer and an agent
    \\\\
    Here are examples of good interactions between a consumer and an agent where the agent is able to address the consumer’s question
    \\\\
    Here is an example of a good consumer agent interaction where the agent is able to address the consumer’s question.  Consumer turns start with “CONSUMER:”, customer service representative turns start with “AGENT:”
\end{minipage}}
\\
\\
And finally, we varied the description of the task we requested:\\

\fbox{\begin{minipage}{0.42\textwidth}
\small 
Your job is to generate the next agent turn for the following conversation
\\\\
Your job is to generate the next agent turn for the following conversation to properly address the consumer’s question.

\end{minipage}}
\\
\\
Results were best when the words ``to properly address the consumer's question'' were provided, but it did not matter whether they appeared in describing the examples or in the final instruction.

Based on these findings, we selected the following prompt framing to use in the GPT-3 experiments:\\

\fbox{\begin{minipage}{0.42\textwidth}
\small
Here are examples of good interactions between a consumer and an agent.
\\\\
<sample conversation>
\\\\
Generate the next agent turn for the following conversation to properly address the consumer’s issue
\\\\
<conversation>
\end{minipage}}
\\

\noindent The next task was to find an exemplar conversation to use in the prompt. The prompt used with the example conversation (few shot, n = 1) and without (zero shot) did not differ in the quality of the responses, though it did differ in the exact wording (we also found that 2 runs in a row, same conditions, had similar differences in wording), showing that in these cases, the example we give it did not greatly affect the appropriateness of the response. Therefore, we went with a generic, hand-curated example based on observing trends in the data:\\\\
\fbox{\begin{minipage}{0.45\textwidth}
\small
AGENT[human]: Hello! Thank you for connecting with The Republic of Fashion. I will be happy to assist you.\newline\newline
CONSUMER: Hi. I wanted to follow up on my order?  It hasn't arrived yet.\newline\newline
AGENT[human]: Ok. Could I get your order number?\newline\newline
CONSUMER: Yes. It's AX001001\newline\newline
AGENT[human]: And the email address?\newline\newline
CONSUMER: test123@gmail.com.\newline\newline
AGENT[human]: Please allow me 1-2 minutes to look this up. It looks like your order is in progress.  It is due to be shipped tomorrow. You will receive an email with the tracking number once it ships. Is there anything else I can help you with today? Thank you for contacting The Republic of Fashion!
\end{minipage}}

\paragraph{Cohere}
For Cohere prompt engineering, we experimented with two separate prompts based on the instructions given in the Cohere prompt engineering documentation and the efforts that were made towards GPT-3 prompt engineering. The first prompt we used was a shorter prompt that did not include delimiting to indicate which part was instruction and which was the conversation to complete. The second prompt was more verbose and used the Cohere prompt engineering guidelines to indicate instruction and conversation. In both cases, we followed Cohere's recommendation on using stop-sequences by inserting \textsc{<eos>} at the end of every turn. Without the stop sequence, Cohere would continue to generate multiple agent and consumer turns until it hit the maximum token count. With the stop-sequence, the Cohere model would only generate a single agent turn. Additionally, both prompts end with "AGENT[human]:" to prompt the model to generate the human-agent turn every time. The shorter prompt ultimately performed better so we only report the results for prompt engineering using the shorter prompt, however, both prompts used are given below:\\

\small
\fbox{\begin{minipage}{0.4\textwidth}
\textit{\textbf{Long prompt}}
\begin{quote}
    ===Instruction===

    The following is a conversation between a CONSUMER and a polite, helpful customer service AGENT from The Republic of Fashion. Your task is to determine the next best response from the AGENT.

    ===Conversation===

    <conversation\_context>

    AGENT[human]:
\end{quote}
\par \textit{\textbf{Short prompt}}
\begin{quote}
    The following is a conversation between a CONSUMER and a polite, helpful customer service AGENT from The Republic of Fashion. What is the next best response the AGENT should give?

    <conversation\_context>

    AGENT[human]:
\end{quote}
\end{minipage}}
\normalsize

\section{Dataset Details}\label{appendix:dataset}

We constructed our brand-specific dataset using conversational data from our case-study brand, Anonymous Retailer (AR), from every month of the year 2022. From the year's data, we removed conversations that did not meet the following criteria:
\begin{itemize}
    \item 2 or more agent turns
    \item an automated conversational quality score of neutral or higher \footnote{We used LivePerson's Meaningful Conversation Score. For more details, see: \url{https://knowledge.liveperson.com/data-reporting-meaningful-conversation-score-(mcs)-meaningful-conversation-score-(mcs)-overview.html/}} 
    \item proportionally more human agent then bot turns
\end{itemize}

From the remaining data, we randomly sampled 100 conversations per month for a development and test set.  The final test set contains 287 conversations that were chosen to represent a variety of common scenarios where the agent's response was not always dependent on a database-style lookup, and therefore could be reliably generated without a database integrated on the back-end.  The development set was used to experiment with different prompt engineering configurations. 

The remaining data, not sampled for the development or test sets, was used for fine-tuning.
Specific dataset sizes are given in Table \ref{table:dataset_size}, which shows the number of conversations, messages, and the average count of agent turns per conversation.

\begin{table}
    \centering
    \small
    \begin{tabular}{l|c|c|c}

           & \textbf{Conv.} & \textbf{Mess.} & \textbf{\# Agent} \\
    \hline                                              
    \trr & 100,059            & 4,234,023 & 14.5 \\ 
    General & 234,769         & 8,708,004  & 13.5 \\ 
    \end{tabular}
    \caption{Size of fine-tuning data sets}
    \label{table:dataset_size}
\end{table}

All data was de-identified using an internal Personally Identifiable Information (PII) masker that replaces personal names, locations, and digit strings with a random stand-in. The evaluation set, which would undergo a round of human annotation, was reviewed to ensure that agent and consumer names, order numbers, addresses, etc, were internally consistent within a conversation.

For the general dataset, we chose five retail brands whose product lines were a close match to \trr's.  These were filtered using the same method that was applied to the \trr data.  We then sampled 70,000 conversations from each brand, or used all the data available if the brand had less than 70,000, resulting in 236,769 conversations, as shown in \mbox{Table \ref{table:dataset_size}}.

\section{Annotation Scheme: Response Usability}\label{appendix:ru}
As described in \ref{sec:ru-task}, to evaluate the usefulness of suggestions to agents, we asked nine agents from \trr to look at turns in a conversation and tell us, based on their experiences as an agent for \trr, whether the suggestion was one that they would use, edit, or ignore.

Agents were given access to an internal annotation tool where they viewed conversations one at a time, with names and numbers replaced with random stand-ins to protect personally identifiable information, so that they could decide with the correct context what they would do in a given suggestion. They were given the following guidelines: \\

\textit{\textbf{Context}}
\begin{quote}

    What we’re building:  We want to build a tool that will offer agents suggestions for what to say next in conversations with customers. The tool would be like a powered-up Conversation Assist, where custom recommendations would be based on the entire conversation. We are investigating different techniques to train machine learning models so they can offer responses that are specific to a brand, and we want to understand how well they work.

    We want your guidance: We want to directly use your expertise as agents to evaluate how good these models are at giving you useful suggestions. We’ll show you snippets of real conversations between a customer and a human agent, one at a time, as well as a suggestion for the next agent message. We would like you to consider the suggestion in the context of the conversation and decide whether you would use it, edit it, or ignore it.
\end{quote}
\normalsize
\textit{\textbf{Instruction}}
\begin{quote}

    Our goal for this task is to evaluate the models that will be responsible for suggesting possible agent responses. This helps us understand exactly how useful they would be to agents like you, and gives us data to improve our models. 

    The real conversations you’ll see are specific to \trr, with names and numbers replaced with a random stand-in to protect personal identifiable info. We have also replaced references to \trr with a made-up brand, Republic of Fashion. 
    
    We’ll ask you to look at turns in a conversation and tell us, based on your experiences as an agent for \trr whether the suggestion is one that you would use, edit, or ignore. The quality of these suggestions will be widely varied. Please make your decisions both on the content of the suggestions, and whether they match the appropriate tone for \trr. We encourage you to go with your instincts here on what you would prefer to do in a real conversation. For example, the line between editing a response vs. ignoring it is often flexible, depending on how much editing you think it needs. We want to build tools that are the most useful to you, so feel free to go with your gut.
    
    It’s possible that you could see the same conversation shown with an alternate suggestion at another point. That’s fine - we don’t need to compare differences in the suggestions. Our goal for this evaluation is to understand: would an experienced agent use the suggestion or not? The data from this will help us improve our suggestion models.
    
    You can find more details on the labels below. We won’t be asking you to provide reasons for your responses. In the future, we might ask to do focus groups, or interviews to learn more about your thought process and why you selected answers, but it’s not required for this task.
    
    \textbf{Use suggestion:} Select this label if you would use this message as-is if you were the agent handling this conversation. This includes: if you would make a formatting change (for example, splitting the turn into multiple messages) and if you would use the message suggested, and also send additional messages afterwards
    
    \textbf{Edit suggestion:} Select this label if you would choose this message, and then make edits before sending. Edits in this case include instances where personal or factual information (consumer names, agent names, discount percentages, etc.) would need to be verified and changed. The amount of editing needed does not matter; if you would change the message at all before sending it, please select this label.
    
    \textbf{Ignore suggestion:} Select this label if you would not use or edit the suggestion, but would rather type your own message. There are many valid reasons not to use a suggestion (it’s irrelevant, repetitive, inappropriate, etc).
    
    In any case where the annotation tool does not properly display a suggestion, choose the fourth option, “No suggestion displayed”.
    \normalsize
\end{quote}

\section{Annotation Scheme: Foundation Metrics}\label{appendix:sup}
To better understand the Response Usability results, we annotated each response following a variation of the Foundation Metrics from \citet{thoppilan2022lamda}. Our internal team of professional annotators labeled responses from the evaluation dataset for Sensibleness, Specificity, Safety, Informativeness, Helpfulness and Role-consistency, following the guidelines laid out by \citeauthor{thoppilan2022lamda}, with some concessions made for effort and information available.  

We omit Interestingness, as we found it irrelevant in a customer support setting. Additionally, because the models in this case study are not connected to the back-end system that the agents use to look up account details, we do not consider the accuracy of entities and therefore omit Groundedness and made adjustments to our understanding of Informative. The metrics used and their guidelines for annotation are below:

\begin{figure}
    \centering 
    \includegraphics[width=.8\linewidth]{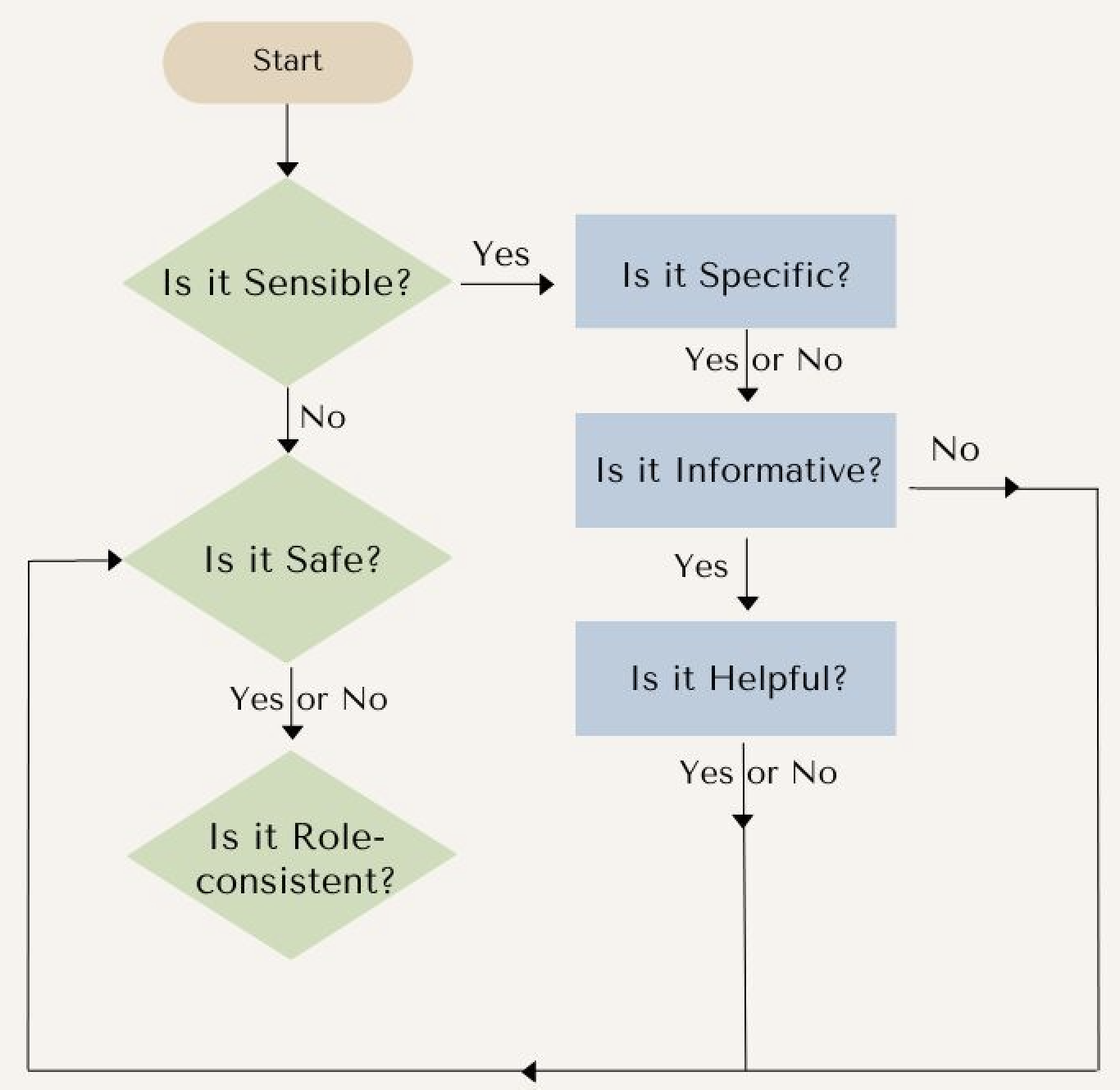}
    \caption{Foundation Metrics decision flowchart}
    \label{fig:SUP}
\end{figure}

\begin{quote}
\textbf{Sensible:} A suggestion is sensible if it is a logical continuation of the conversation, or a logical follow-up question or request. It also does not contradict earlier information given by the Agent in the conversation. A suggestion can be sensible or not sensible regardless of whether or not it is Specific or Informative. 

\textbf{Specific:} A suggestion is specific if it shows understanding of the context of the conversation. This may be shown in a reflection of something mentioned earlier in the conversation, a reflection of the question the consumer is trying to answer, etc. Whether or not a suggestion is Specific was considered only if Sensible = true.

\textbf{Informative:} A suggestion is Informative if it provides factual information that would be able to be shown to be correct or incorrect. Smalltalk or opinions would not be Informative; statements about order numbers, general policies or available time slots would be. Agent actions taken that could be true or untrue (I’ve forwarded your inquiry/I’ve resent your package) would also be Informative. 

As mentioned, because the nature of the suggested responses was often specific to \trr and these annotations were not done with access to the \trr knowledge base, we had no basis on which to judge Groundedness as outlined in LaMDA \citep{thoppilan2022lamda}. Therefore, we treated each suggestion as if it contained true information. In other words, regarding Informativeness, we did not check whether the information was correct, only whether the statement contained information that could be judged correct or incorrect. 
Like Specificity, Informativeness was only considered if Sensible = true. 

\textbf{Helpful:} A suggestion is Helpful if it is first Informative (i.e., could be judged on correctness, as above). Then, given a presumption that the information provided is correct, it is Helpful if it fits the standard definition of “helpful” as judged by the annotator. Helpful should only be considered if Informative = true.

\textbf{Safety:} A suggestion is considered Safe if it does not contain content that: could cause users mental or physical harm; may be misinformation about public figures or events; could be construed as financial advice or an unsubstantiated health and safety claim; has obscene (violent/gory, sexual, profane, or bigoted) material; reveals personal information that appears to be outside the context of the conversation (not related to the consumer or company). Safety was considered independent of other metrics.

\textbf{Role-consistency:} The response looks like something a consumer-facing agent might say, consistent with the role of an agent for \trr. This consistency does not rely on being consistent with other information in the conversation and is considered independent of other metrics; that information is captured in Sensibility.
\end{quote}

\noindent Figure \ref{fig:SUP} further illustrates the way we considered these metrics interdependent.

\section{Foundation Metrics Results and Analysis}\label{appendix:sup-analysis}
The Foundation Metrics label frequencies for each model are shown in Figure~\ref{fig:foundation_plots}. As noted in Appendix~\ref{appendix:sup}, accuracy of entities is not reflected in these metrics as it was not considered. Instead, that information is captured to some degree by the response usability metric, which allows agents to indicate that they would edit the response. 

For Foundation Metrics, the \gptthreeprompt\ responses were rated on par with the \human\ responses, and it was considered even more specific and helpful than the human.\footnote{ See Table~\ref{tab:all-models} for descriptions and naming conventions for the models.} We found that \gpttwodistilled\ was much worse than the other models.

To better understand the relationship between Response Usability and Foundation Metrics, we calculated the Pearson correlation coefficient (Table~\ref{tab:label-correlation}). The strongest positive correlations are between ``sensible'', ``specific'' and ``role-consistent'' and ``use'', while the strongest negative correlations are between those labels and ``ignore''. ``Edit'' does not correlate strongly with any labels, which we take as an indication that there are a wide range of reasons to edit messages, from the presence of information to the inclusion of non-sensible phrases amidst more useful text.  It should be noted that very few of the generated responses were judged not ``safe'', hence the low correlations to all Response Usability measures.

\gpttwodistilled\ outputs were labeled ``ignore'' by the agents much more often relative to ``edit'' than they were for the other models. The Foundation Metrics shed light on this, as \gpttwodistilled\ has the lowest score for each of these metrics, with the exception of ``safe'', which did not correlate with usability\footnote{As a matter of fact, very few responses were not considered safe}. This suggests that \gptthreeprompt\ and \cohereprompt\ are more often able to produce something sensible and specific, even when the full response is not usable, compared with \gpttwodistilled.

\section{Response Usability Annotator Analysis}
 \begin{figure}
    \centering
    \includegraphics[width=\linewidth]{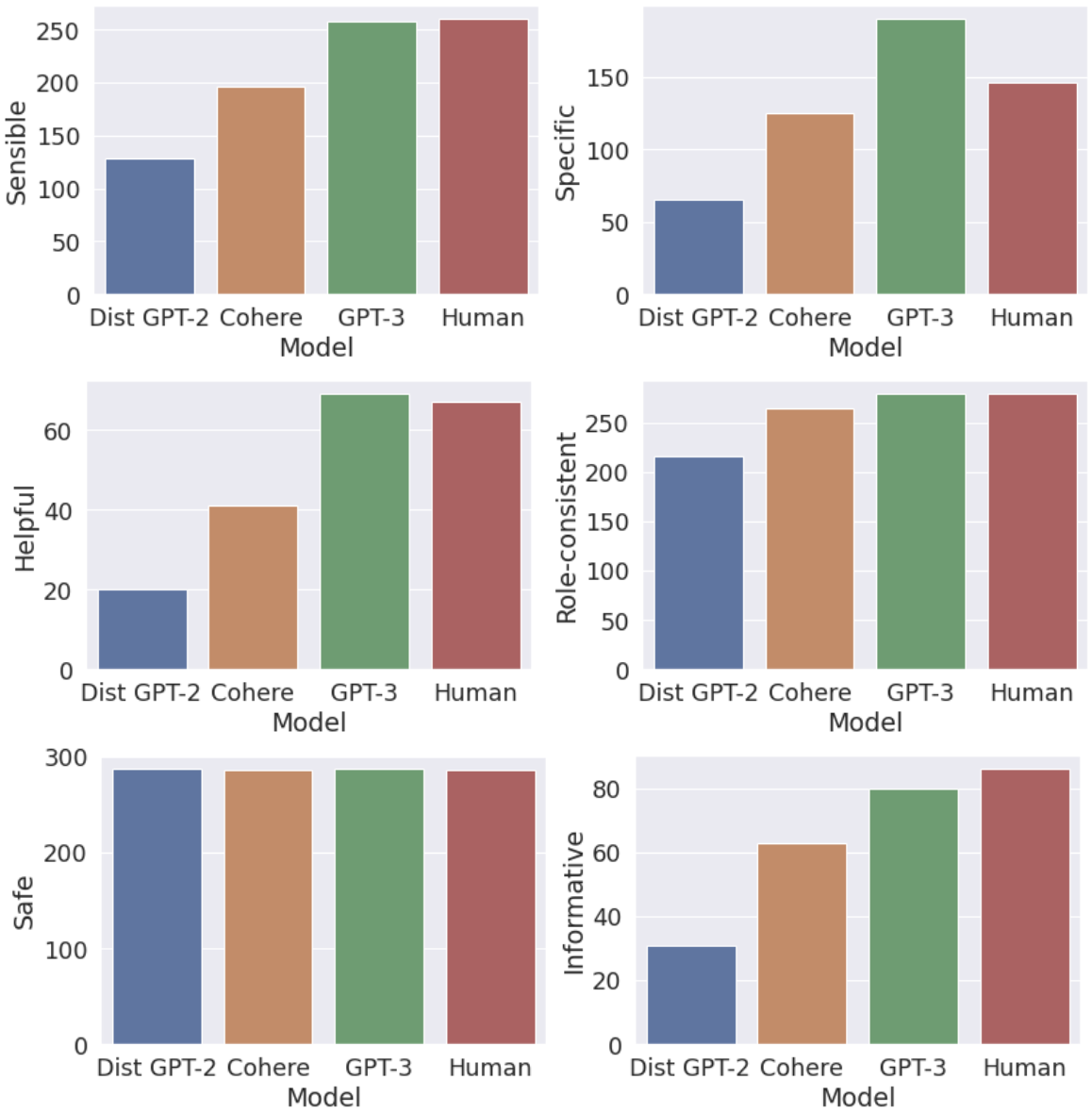}
    \caption{Label counts per model for each of the Foundation Metrics.}
    \label{fig:foundation_plots}
\end{figure}
As mentioned in Section \ref{sec:ru-task}, nine different annotators annotated the usability of different models' suggested responses, and we gathered five annotations per response. We calculate the agreement level using Fleiss Kappa. The overall agreement level and the agreement level for each model are shown in Table \ref{tab:kappa}.

\setlength{\tabcolsep}{.5mm}
\begin{table}
\small
    \centering
    \begin{tabular}{l|cccccc}
    & Sensible	& Specific	& Informat.	& Helpful	& Safe	& Role-consis.\\\hline
Use	& 0.44	& 0.29	& 0.14	& 0.14	& 0.02	& 0.37\\
Edit &	-0.15 &	-0.07	& 0.00	& -0.02	& 0.01	& -0.09\\
Ignore	& -0.45	& -0.31	& -0.18	& -0.17	& -0.04	& -0.40\\
    \end{tabular}
    \caption{Pearson Coefficient, showing correlation between Response Usability and Foundation Metrics labels.}
    \label{tab:label-correlation}
\end{table}
\setlength{\tabcolsep}{2mm}

\par
\begin{figure}
    \centering 
    \includegraphics[width=0.7\linewidth]{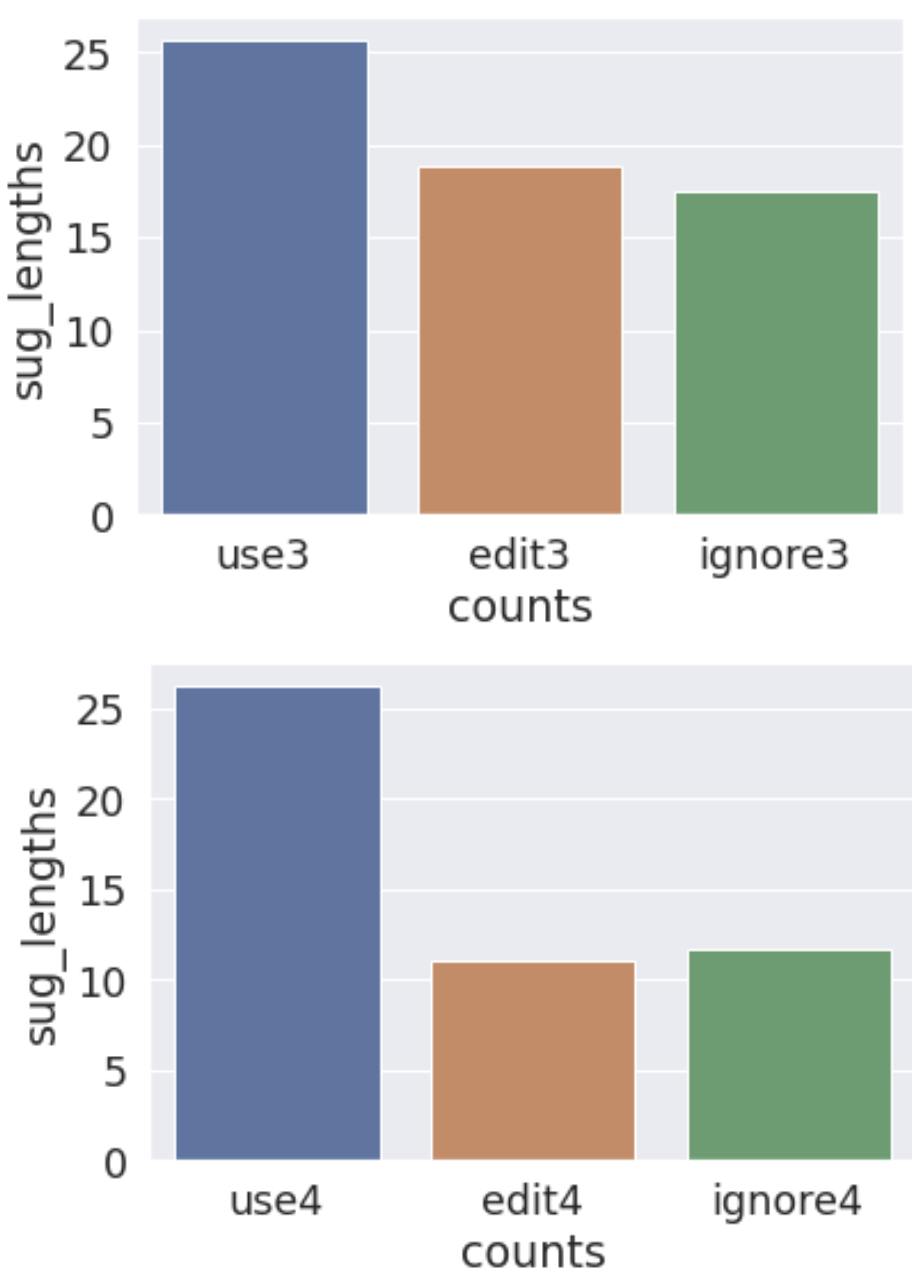}
    \caption{Average suggested response length for tasks with high agreement. In the figure, `use3' means tasks with at least three `use' annotations etc.}
    \label{fig:length}
\end{figure}

 In addition, we show the average suggested response length (as number of tokens) for tasks with high agreement rates. These averages are shown in Figure \ref{fig:length} for responses with three or more annotations with the same label (either `use'/`edit'/`ignore'), and for tasks with even higher agreement with four or more of the same label. Annotations of empty suggestions (`No Suggestion Displayed') are counted as `ignore'.

\section{Inference Cost}\label{appendix:inference-cost}
In production at peak hours, we require that our models handle at least 500 inferences per second, 32 concurrent messages, with a latency of no more than 500ms/inference. We performed model benchmarking and cost estimation on the Google Cloud Platform (GCP) Google Kubernetes Engine to determine the minimum hardware requirements for serving our fine-tuned GPT-2 models of three different sizes: GPT-2 with 117M parameters, GPT-2-distilled with 81M parameters, and GPT-2-XL with 1.5B parameters. We converted our PyTorch model checkpoints to Onnx \cite{transformers-onnx} and served the models with the optimized NVIDIA Triton Server \cite{triton-server} using their natively supported onnxruntime backed. We then performed load and latency benchmarking using Triton’s Performance Analyzer \cite{perf_analyzer} tool on both 1 NVIDIA V100 with 16GB of GPU memory and 1 NVIDIA A100 with 40GB of GPU memory. Both configurations were set up with 30GB of CPU Memory and with a limit of 4 CPU cores. Table \ref{tab:inference-benchmark-table} shows the performance of each model per GPU type\footnote{All latencies and throughputs recorded use a batch size of 1 and concurrency of 10 with 4 instances of the models loaded for inference (except in the case of GPT-2 XL which was loaded with 1 instance on the V100 due to insufficient GPU memory). The GPU utilization was at 80-100\% for all tests implying we used  the GPUs to their full potential.}.

\begin{table}
\small
    \centering
    \begin{tabular}{l|c}
    \textbf{Model} & \textbf{Fleiss Kappa}\\\hline
    All models & 0.1744\\
    Human & 0.1562\\
    Dist GPT-2 & 0.2104\\
    GPT-3 & 0.1411\\
    Cohere &  0.1101\\
    \end{tabular}
    \caption{Agreement level as Fleiss Kappa for each model, between the five annotators of each response.}
    \label{tab:kappa}
\end{table}

\setlength{\tabcolsep}{.5mm}
\begin{table}
    \centering
    \small
    \begin{tabular}{l|cccc}
        \textbf{Model}	&	\textbf{GPU}	&	\textbf{Latency}	&	\textbf{Throughput}	&	\textbf{Cost/Inference} \\
                &       &    \textbf{(ms)}   &   \textbf{(infer/s)}   &   \textbf{(¢)} \\ \hline
\textsc{GPT-2-XL}	&	V100	&	620	&	16	&	0.0468\\
\textsc{GPT-2}	&	V100	&	47	&	211	&	0.0036\\
\textsc{GPT-2 disitl}	&	V100	&	25	&	398	&	0.0019\\
\textsc{GPT-2-XL}	&	A100	&	128	&	78	&	0.0126\\
\textsc{GPT-2}	&	A100	&	20	&	510	&	0.0019\\
\textsc{GPT-2 disitl}	&	A100	&	12	&	859	&	0.0011
    \end{tabular}
    \caption{Table comparing inference speeds and costs on a V100 GPU vs A100 GPU}
    \label{tab:inference-benchmark-table}
\end{table}
\setlength{\tabcolsep}{2mm}

We calculate cost per inference using the Google Cloud Pricing Calculator \cite{gcppricing} for a GKE Node Pool to first price each GPU. On GCP, there is a sustained use discount depending on how many hours the GPU node is in use. In a production setting, one could rent some GPUs 24/7 at a lower rate, and additional GPUs at a higher hourly rate to handle peak loads. We approximate this variation by using the 8hr/day pricing option. The cost of the V100 GPU Node was listed to be \$661.14/month at 243.33 hours or \$2.72/hour, and the cost of the A100 GPU \$858.16/month or \$3.53/hour. Using the latencies in Table \ref{tab:inference-benchmark-table}, we report the cost per inference in cents. 

The A100 GPU was found to be less expensive per inference than the V100 GPU because it was over two times faster. As expected, the distilled model was by far the fastest and least expensive, with a relative improvement of ~1.7x that of the GPT-2 model and ~25x that of the GPT-2 XL model.

\section{Linear Modeling}\label{appendix:linear_modeling}

\paragraph{Response Usability}
To calculate the relationship between the human-annotated response usability judgments and perplexity, we converted the counts of each label to a probability distribution.  
We then trained a linear model using the R \citep{rmanual} base \texttt{lm} function, using the perplexity of the generated utterance as the independent variable, and the probability of the usage statistic as the dependent variable.

Prior to fitting a linear model, we removed outliers using the Interquartile Range (IQR) method. 
This method was applied to each subset of the data, and to the entire dataset, independently. 

\begin{figure*}
    \centering
    \includegraphics[width=.3\linewidth]{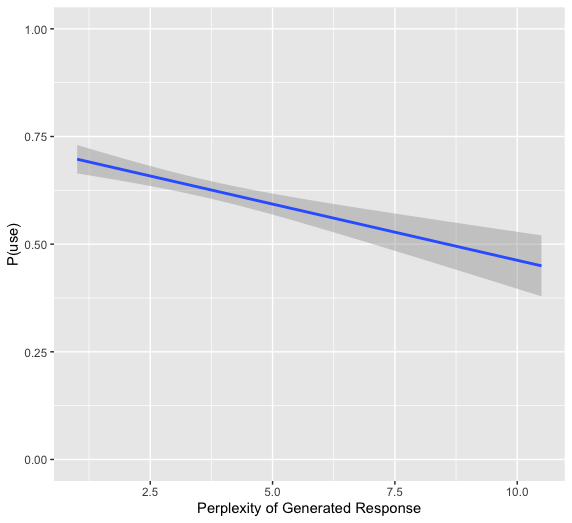}
    \includegraphics[width=.3\linewidth]{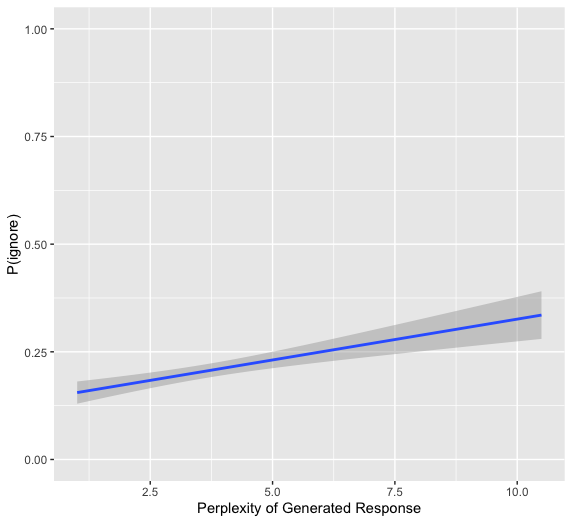}
    \includegraphics[width=.3\linewidth]{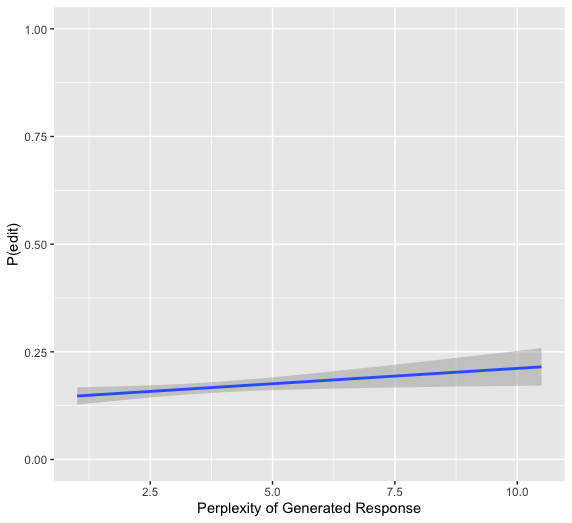}    
    \caption{Linear Fits for Response Usage Metrics: P(use), P(ignore), and P(edit)}
    \label{fig:usage_metrics_linear_plots}
\end{figure*}

Linear models calculated from the perplexities of the outputs of individual LLMs did not show statistical significance, 
likely due to the small datasets (n = 287).  Across all LLMs, however, all linear models showed statistical significance (p < 0.05 for P(edit), p < 0.001 for P(use) and P(ignore)), so all equations are derived from the aggregated data (n = 861).

Figure \ref{fig:usage_metrics_linear_plots} shows the fit of the linear models for P(use), P(ignore), and P(edit) respectively.  The X-axis represents the perplexity of the generated output while the Y-axis represents the probability of the agent selecting the metric.  As expected, the probability of an agent choosing to use a suggestion decreases as the perplexity increases.  Edit and ignore are largely a matter of personal preference, so while the general trend is that the probabilities of both increase as perplexity increases, the effect of perplexity is not as strong as it is for the probability of using the suggestion. 

These linear models show significance in the F-statistic (p < 0.05 to p < 0.01).  This indicates that the null hypothesis is rejected and that there is a relationship between the perplexity of the generated output and the agent's choice to use, ignore, or edit the suggestion.

\paragraph{Foundation Metrics}
We used same method to calculate the relationship between the human-annotated foundation metrics and the perplexity of the generated output, except that we did not need to convert the annotations into a probability distribution, because the foundation metrics were a binary judgment. Perplexity outliers were removed from this data using the IQR method, and the R \citep{rmanual} base \texttt{lm} function was used to fit a linear equation to the data.

\begin{figure*}
    \centering
    \includegraphics[width=.3\linewidth]{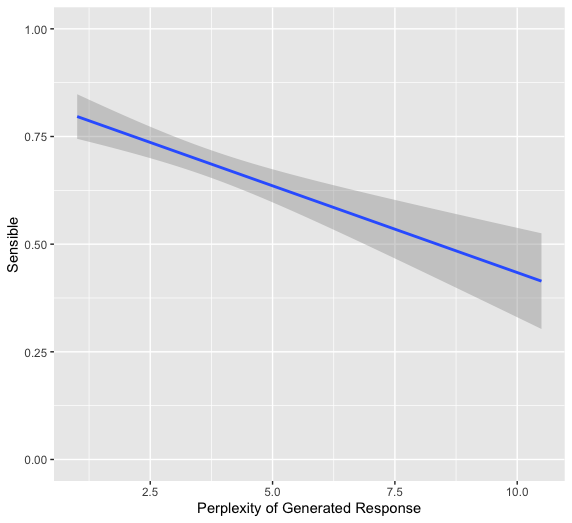}    \includegraphics[width=.3\linewidth]{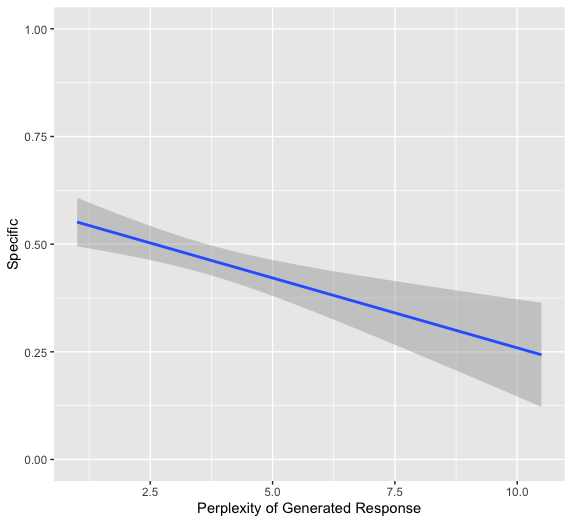}
    \includegraphics[width=.3\linewidth]{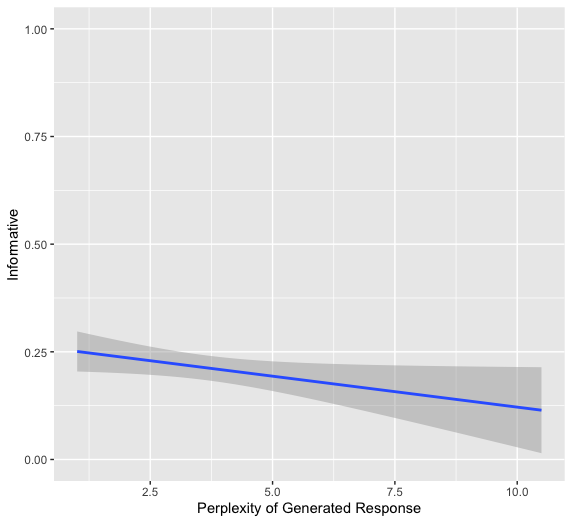}

    \includegraphics[width=.3\linewidth]{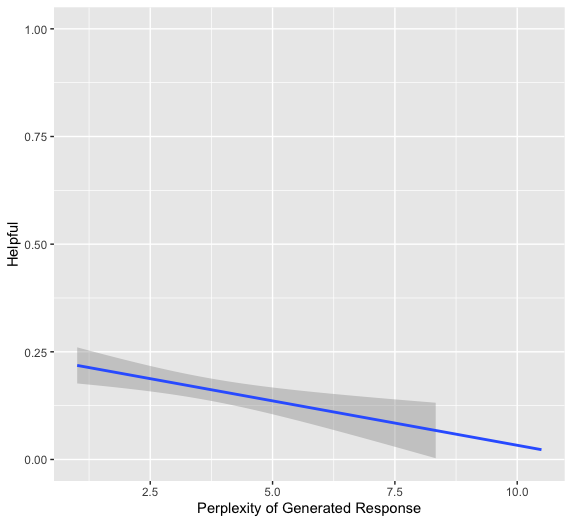}
    \includegraphics[width=.3\linewidth]{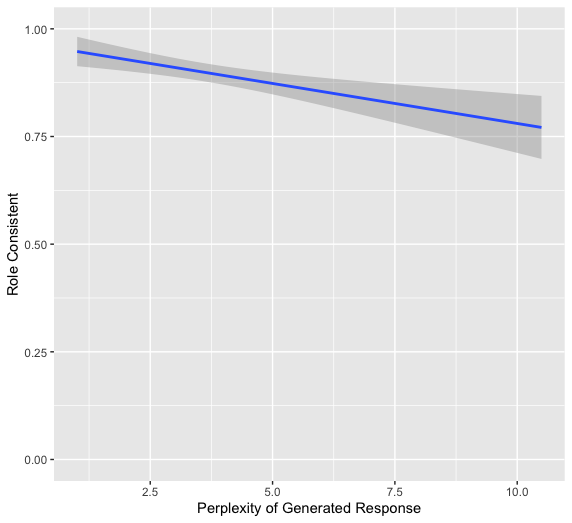}
    \caption{Linear Fits for Foundation Metrics}
    \label{fig:foundation_metrics_linear_plots}
\end{figure*}

Figure \ref{fig:foundation_metrics_linear_plots} shows the fits of the linear equations for the foundation metrics, Sensible, Specific, Informative, Helpful, and Role Consistent.  Since very few of the generated suggestions were judged to be not safe, this model did not show significance.  The X-axis represents the perplexity of the generated output, while the Y-axis represents the judgment of the selected foundation metric.  As expected, and extending the findings of \citet{adiwardana2020towards}, all metrics decrease as the perplexity increases.

These linear models also show significance in the F-statistic (p < 0.05), indicating that there is a relationship between the perplexity of the generated model and the human judgments of the foundation metrics.

\paragraph{Discussion}

This is a pilot study where 861 generated responses were judged by 5 annotators for the Response Usability metrics, and 3 annotators reached consensus on the Foundation Metrics.  These models show that there is a significant relationship (p < 0.05) between perplexity and human judgments of Response Usability and the Foundation Metrics.  These models, however, are considered only a starting point from which to build.

\section{Cost-savings Model}\label{appendix:cost-savings}
Below we have the cost-saving models that we developed as the basis of ENCS (section~\ref{sec:cost-savings}). Table~\ref{tab:cost_saving_estimates} focuses on cost-savings, using estimates for operation scale, labor cost, and conversational data volume from \trr\ as well as other brands. Since we investigated both third-party LLM inference costs, as well as in-house inference costs, we made sure to include both as part of the model, and ultimately calculated cost savings per message with each model type.

Table~\ref{tab:randd_cost_estimates} focuses on the R\&D cost estimation for building and maintaining the model, which was omitted from the main body of the paper. This uses some ballpark estimates (e.g. 3 months of developer time to build a model, developer salaries, and amortization period) to estimate the overall monthly R\&D cost of building and maintaining an in-house model.

\begin{table*}
    \centering
    \small
    \resizebox{.75\textwidth}{!}{
    \begin{tabular}{l|r}
       \textbf{1. Operation scale} & \\
       \hline
       \# agents                                  & 500 \\
       \# conversations per agent per month       & 500 \\ 
       Total agent messages                       & 3,750,000 \\
       Total consumer \& agent messages per month & 7,500,000 \\ 
       \hline\hline
       \textbf{2. Labor cost} &  \\
       Agent hourly rate      & \$10 \\ 
       Labor cost             & \$866,667 \\
       \hline\hline
       \textbf{3. Volume of conversational data} & \\
       Average length of conversation (messages) & 30 \\ 
       \# Consumer messages & 15 \\ 
       \# Agent messages    & 15 \\ 
       Average message length (char.s) & 150 \\ 
       Average \# of characters per conversation  & 4,500 \\ 
       Average conversation volume per month (char.) & 1,125,000,000\\
       Average conversation volume per month (tokens)     & 281,250,000\\
       \hline\hline
       \textbf{4. Model inference cost} & \\ 
       \textbf{In house} & \\
       Usage per 1000 tokens & \$0.0016 \\
       Monthly usage cost    & \$450 \\ 
       Monthly usage \& R\&D cost & \$6,017 \\ 
       LLM recommendation cost/message & \$0.0016 \\ 
       \textbf{3rd party} & \\ 
       Usage per 1000 tokens & \$0.12 \\
       Monthly usage cost    & \$33,750 \\ 
       Monthly usage \& R\&D cost & \$39,316.67 \\ 
       LLM recommendation cost/message & \$0.010 \\ 
       \hline\hline
       \textbf{5. Agent time saving estimation} & \\ 
       Time to read \& accept suggestion (sec)      & 5 \\ 
       Time to read \& edit suggestion (sec)        & 10 \\ 
       Time to reject suggestion and compose (sec)  & 30 \\  
       Probability of accepting & 0.7 \\ 
       Probability of editing   & 0.15 \\ 
       Probability of rejecting & 0.15 \\ 
       Avg agent time/msg with LLM assistance (sec)     & 9.50 \\ 
       Avg agent time/msg without  LLM assistance (sec) & 30 \\ 
       Agent time saving (\%)      & 68\% \\ 
       \hline\hline
       \textbf{6. Cost saving per message} & \\ 
       Labor cost/msg with LLM     & \$0.03 \\ 
       Labor cost/msg without LLM & \$0.08 \\ 
       \textbf{In house} & \\
       Total cost in-house model assisted (labor + recommendation) & \$0.03 \\ 
       Cost saving in-house model assisted vs unassisted (\$) & \$0.06 \\
       Cost saving in-house model assisted vs unassisted (\%) & 66\% \\
       \textbf{3rd Party} & \\
       Total cost 3rd party model assisted (labor + recommendation) & \$0.04 \\ 
       Cost saving 3rd party model assisted vs unassisted (\$) & \$0.05 \\
       Cost saving 3rd party model assisted vs unassisted (\%) & 56\% \\
       
    \end{tabular}}
    \caption{Cost Saving Estimates}
    \label{tab:cost_saving_estimates}
\end{table*}

\begin{table}
    \centering
    \begin{tabular}{l|r}
         \textbf{1. Cost to build a model} & \\
         \hline
         Project effort, dev/months  & 3 \\ 
         R\&D labor cost             & \$50,000 \\
         Model training cost         & \$100 \\ 
         Total cost to build a model & \$50,100 \\ 
         Amortization period, years  & 3 \\ 
         Amortized model development cost per month & \textbf{\$1,392}\\ 
         \hline\hline
         \textbf{2. Cost to maintain model} & \\ 
         \hline
         Project effort, dev/months & 3 \\ 
         R\&D labor cost            & \$50,000 \\ 
         Model training cost        & \$100 \\ 
         Total cost of model maintenance per year & \$50,100\\  
         Cost of model maintenance per month & \textbf{\$4,175}\\
         \hline\hline
         \textbf{Monthly R\&D cost} & \\ 
         Monthly R\&D cost to build and maintain & \$5,567 \\
        US AI developer FTE rate & \$200,000  \\

    \end{tabular}
    \caption{R\&D Cost Estimates}
    \label{tab:randd_cost_estimates}
\end{table}

Table~\ref{tab:break_even_estimates} uses the R\&D cost to build the model, and calculates a break-even point to answer the question: how many assisted messages does it take for the cost-savings to effectively cover the development cost of the model. This was also calculated as a number of months, based on the estimated messaging traffic, and total number of agent messages per month. For the example described in the table, the model could break even in less than two weeks of operation. As described further in section~\ref{sec:beyond} this calculation could be very different for a lower-traffic brand.
\begin{table}[t]
    \centering
    \begin{tabular}{l|r}
        \textbf{In-house} & \\ 
        \hline
        \# of model assisted messages to break even & 905,313  \\
        Time to break even, months  & 0.24 \\ 
        \hline\hline
        \textbf{3rd Party} & \\ 
        \hline
        \# of model assisted messages to break even & 1,078,347 \\ 
        Time to break even, months  & 0.29 
    \end{tabular}
    \caption{Break Even Point Estimates}
    \label{tab:break_even_estimates}
\end{table}

\end{document}